\pdfoutput=1
\documentclass{article}
\usepackage[T1]{fontenc}
\usepackage{t2spo_preprint}
\usepackage[charter]{newtxmath}
\usepackage{amsmath,amsfonts,bm}
\usepackage{url}
\usepackage{booktabs}
\usepackage{colortbl}
\usepackage{graphicx}
\usepackage{algorithm}
\usepackage{algpseudocode}
\usepackage{hyperref}

\title{T2SPO: Trajectory-to-Step Policy Optimization for Agentic Reinforcement Learning}
\newcommand{\PreprintAuthorNames}{Bo-Wen Zhang, Junwei He, Maoqi Liu, Feiran Li, Song-Lin Lv, Wentao Ma, Rongyi Lin, Shuhan Zhong, Lan-Zhe Guo}
\author{%
\begin{minipage}[t]{\dimexpr\textwidth-2\tabcolsep\relax}
\centering
Bo-Wen Zhang\textsuperscript{1,2,3,*\ensuremath{\dagger}}\quad
Junwei He\textsuperscript{3,*}\quad
Maoqi Liu\textsuperscript{3}\quad
Feiran Li\textsuperscript{3}\\[3pt]
Song-Lin Lv\textsuperscript{1,2}\quad
Wentao Ma\textsuperscript{3}\quad
Rongyi Lin\textsuperscript{3,\ensuremath{\ddagger}}\quad
Shuhan Zhong\textsuperscript{3}\quad
Lan-Zhe Guo\textsuperscript{1,2,\S}\\[6pt]
{\normalfont\normalsize
\textsuperscript{1}State Key Laboratory of Novel Software Technology, Nanjing University\\
\textsuperscript{2}School of Intelligence Science and Technology, Nanjing University\\
\textsuperscript{3}ByteDance\\[3pt]
{\footnotesize\texttt{guolz@nju.edu.cn}}}
\end{minipage}%
}
\newcommand{\PreprintAuthorNotes}{%
\begingroup
\renewcommand{\thefootnote}{}%
\footnotetext[0]{\textsuperscript{*}Equal contribution.
\textsuperscript{\ensuremath{\dagger}}Work done during internship at ByteDance.
\textsuperscript{\ensuremath{\ddagger}}Project lead.
\textsuperscript{\S}Corresponding author.}%
\endgroup
}

\hypersetup{
    colorlinks=true,
    linkcolor=PreprintLink,
    citecolor=PreprintLink,
    urlcolor=PreprintLink,
    pdftitle={T2SPO: Trajectory-to-Step Policy Optimization for Agentic Reinforcement Learning},
    pdfauthor={\PreprintAuthorNames},
    pdfsubject={Trajectory-derived step credit for language-agent reinforcement learning},
    pdfkeywords={LLM agents, reinforcement learning, credit assignment, in-context learning}
}

\begin{document}
\maketitle
\pagestyle{plain}
\PreprintAuthorNotes
\raggedbottom

\begin{abstract}
Reinforcement learning enables large language model (LLM) agents to learn multi-step behaviors through interaction with their environments.
However, rewards in many interactive tasks reflect only the final outcome, providing limited guidance on which intermediate decisions advance the task.
Successful training trajectories contain intermediate states that can provide supervision for subsequent interactions.
We introduce \textbf{Trajectory-to-Step Policy Optimization (T2SPO)}, a method that uses past interaction trajectories to provide step-level feedback for policy learning.
T2SPO derives remaining-distance targets from successful trajectories and pairs them with representations of the states visited along the way.
Conditioned on these examples, a pretrained TabPFN regressor estimates the remaining distance to success at each state of a new rollout.
Changes in this distance estimate across consecutive states yield auxiliary credit for agent steps alongside task-level supervision.
As training proceeds, newly completed trajectories refresh the estimator's context, incorporating new experience without updating its parameters.
Experiments with 1.5B and 7B language models on ALFWorld and WebShop show that T2SPO consistently improves overall task success over GRPO.
\end{abstract}

\begingroup
\setlength{\intextsep}{6pt}
\begin{figure}[H]
    \centering
    \captionsetup{font=small}
    \includegraphics[width=0.86\linewidth]{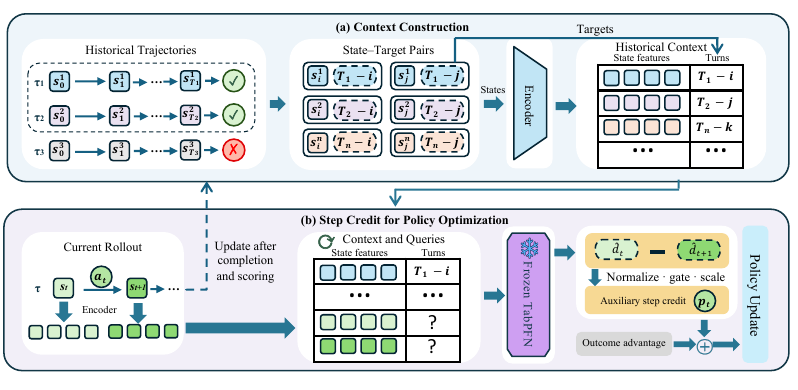}
    \caption{\textbf{T2SPO converts historical trajectories into step-level credit.}
    \textbf{(a) Context construction.} Successful historical trajectories pair each visited state with its own observed number of turns to completion (Section~\ref{sec:state_targets}).
    State representations and these targets form a labeled context.
    \textbf{(b) Step credit for policy optimization.} A frozen TabPFN predicts the remaining distance at adjacent states of a current rollout using the same historical context.
    Their difference is normalized, gated, and scaled into auxiliary step credit, which augments the outcome advantage in GRPO.
    Current trajectories refresh the context only after completion and scoring.}
    \label{fig:t2spo_overview}
\end{figure}
\endgroup

\clearpage
\setlength{\parskip}{6pt plus 1pt minus 1pt}
\section{Introduction}
\label{sec:introduction}
Reinforcement learning enables LLM agents to improve through environment interaction~\citep{ArCHer2024,RAGEN2025}.
In multi-step tasks, the final outcome evaluates the whole trajectory but does not reveal how individual decisions contribute to success.
\textbf{Estimating task progress and its changes across steps} helps connect individual decisions to task outcomes, providing fine-grained signals for credit assignment during policy optimization~\citep{ArCHer2024,TurnPPO2025}.

\looseness=-1
PPO uses a learned critic to estimate expected returns from intermediate states~\citep{PPO2017}.
Combined with generalized advantage estimation, these predictions provide state-dependent training signals for individual actions~\citep{GAE2015}.
This requires training a value model alongside the policy.
GRPO avoids training a critic by estimating its baseline from the rewards of trajectories sampled for the same task~\citep{DeepSeekMathGRPO2024}.
Under outcome-only supervision, all steps within a trajectory share the same outcome advantage, even when some decisions advance the task and others are redundant~\citep{GiGPO2025}.
If all trajectories in a group receive the same outcome reward, these advantages are zero.
This motivates our central question: \textbf{Can historical experience supply fine-grained step credit without training an additional value model?}

Successful trajectories collected during policy training provide supervision beyond their final outcomes: each visited state can be paired with the number of turns remaining along its own successful continuation.
These observed lengths serve as a proxy for task progress, providing supervision for estimating remaining distance at states encountered in later rollouts.
Changes in the predicted distance across consecutive states can then supply auxiliary credit for individual agent steps.

In this paper, we introduce \textbf{Trajectory-to-Step Policy Optimization (T2SPO)}, which implements this approach with a frozen TabPFN regressor~\citep{PFN2021,TabPFNv2Nature2025}.
For each rollout batch, the regressor predicts remaining distance using a shared context of historical state representations and completion-length labels.
Adjacent-state predictions define a potential difference that provides auxiliary step credit~\citep{PotentialBasedRewardShaping1999}.
Newly completed trajectories update the context after scoring, allowing subsequent estimates to incorporate new experience while the encoder and regressor remain frozen.
The credit augments outcome-based advantages in our main implementation and can also enter an actor--critic optimizer as an auxiliary reward (Figure~\ref{fig:t2spo_overview}).

Our contributions are:
\begin{itemize}
    \item We propose \textbf{T2SPO}, a framework that reuses historical successful trajectories to supply auxiliary credit for individual agent steps while retaining task-level supervision.
    \item We develop a progress estimator with a frozen TabPFN regressor and online context updates, yielding auxiliary step credit without training an additional scoring model.
    \item Experiments on ALFWorld and WebShop show higher overall success than the reported GRPO baselines at both scales~\citep{ALFWorld2020,WebShop2022}. Offline evaluations assess distance prediction across context sizes and SVD dimensions.
\end{itemize}

\section{Related Work}
\label{sec:related_work}

\looseness=-1
\noindent\textbf{Reinforcement learning for LLM agents.}
Reinforcement learning is widely used in language-model post-training, with feedback drawn from human preferences or verifiable task outcomes~\citep{InstructGPT2022,DeepSeekR12025,DAPO2025}.
For LLM agents, RL optimizes policies over multiple rounds of interaction, where earlier actions influence later decisions and eventual task outcomes~\citep{ArCHer2024,RAGEN2025,SearchR12025}.
Sparse or delayed feedback makes it difficult to determine which actions contribute to success, motivating methods that connect outcomes to individual steps~\citep{GiGPO2025}.

\looseness=-1
\noindent\textbf{Step-level feedback and credit assignment.}
Step-level feedback can be obtained by training a model to evaluate intermediate states or actions.
For mathematical reasoning, process reward models learn from human step annotations or automatically generated supervision: Let's Verify Step by Step uses human judgments, while Math-Shepherd derives labels from sampled continuations~\citep{LetsVerifyStepbyStep2023,MathShepherd2024}.
For LLM agents, ArCHer and Turn-PPO learn critics that estimate expected returns at the turn level and support policy updates~\citep{ArCHer2024,TurnPPO2025}.
Another approach compares returns following different actions from a shared state.
GiGPO identifies repeated environment states in collected trajectories and groups the corresponding actions to compute relative advantages~\citep{GiGPO2025}.
ARPO instead uses entropy to guide branching during rollout collection and assigns advantages according to the shared and branched portions of the resulting trajectories~\citep{ARPO2026}.
Neither method trains an auxiliary scoring model.

\noindent\textbf{Tabular in-context learning for value estimation.}
Prior-data fitted networks learn to make supervised predictions from labeled context examples by pretraining over synthetic tasks~\citep{PFN2021}.
Tabular foundation models extend this approach to classification and regression, with TabPFN emphasizing prediction from small datasets and TabICL scaling in-context classification to larger tables~\citep{TabPFNv12022,TabPFNv2Nature2025,TabICL2025}.
Closely related applications are $V_0$ and $V_{0.5}$, which combine semantic representations with TabPFN-based prediction conditioned on prior instruction--performance pairs~\citep{V02026,V052026}.
$V_0$ predicts success probability at the initial prompt to allocate rollout budgets and route instructions, while $V_{0.5}$ combines such predictions with sparse on-policy outcomes for baseline estimation and adaptive sampling.
T2SPO instead conditions a frozen TabPFN regressor on representations of intermediate states paired with the remaining lengths of their successful source trajectories.
Differences in predicted remaining distance between adjacent states supply auxiliary step credit for policy optimization.

\section{Preliminaries}
\label{sec:preliminaries}
We first define the interaction and optimization setting, then introduce the regression interface used to transfer trajectory-derived supervision to current states.

\subsection{Multi-Turn Reinforcement Learning}
\label{sec:agent_rl}
\looseness=-1
LLM agents interact with their environments by selecting actions conditioned on observations and prior context~\citep{ReAct2023}.
We use \emph{step}, or \emph{turn}, to denote one complete agent--environment interaction.
Given an instruction $q$, let $s_t$ denote the context available to the policy before turn $t$, typically consisting of the instruction, the current observation $o_t$, and processed interaction history~\citep{dong2026longhorizon}.
The policy $\pi_\theta(a_t\mid s_t)$ generates a textual action $a_t$ autoregressively, after which the environment returns reward $r_t$ and observation $o_{t+1}$.
A trajectory $\tau$ comprises $T$ such transitions, with sparse task feedback provided by the outcome at the end of the episode.

\noindent\textbf{Advantage estimation.}
Outcome-supervised GRPO samples a group of $G>1$ trajectories for the same task and normalizes their final task scores $R_i$~\citep{DeepSeekMathGRPO2024}:
\begin{equation}
    \hat A_i^{\mathrm{out}}
    = \frac{R_i - \overline R}{\sigma_R + \varepsilon},
    \label{eq:grpo_outcome}
\end{equation}
where $\overline R$ and $\sigma_R$ denote the group mean and standard deviation, and $\varepsilon>0$ ensures numerical stability.
This estimator assigns the same advantage to all generated decisions in trajectory $i$ without a learned critic.
For example, an above-average outcome reinforces both a redundant search and the successful purchase that follows it.
When all group outcomes are equal, the outcome advantages are zero even if the trajectories differ in intermediate progress or redundant behavior.

\noindent\textbf{Learned critics.}
PPO can obtain advantages that vary across turns through a learned value function and generalized advantage estimation (GAE)~\citep{PPO2017,GAE2015}.
At a nonterminal turn with zero environment reward, its temporal-difference residual is $\delta_t=\gamma V_\psi(s_{t+1})-V_\psi(s_t)$, where $\gamma\in(0,1]$ is the discount factor.
These values let task outcomes inform updates to earlier decisions.

\noindent\textbf{Policy updates.}
We use the clipped surrogate~\citep{PPO2017,DeepSeekMathGRPO2024}
\begin{equation}
    \ell_{\mathrm{clip}}(\rho, \hat A)
    = \min\bigl\{
        \rho\hat A,
        \operatorname{clip}_{[1-\epsilon,\,1+\epsilon]}(\rho)\hat A
      \bigr\},
    \label{eq:clipped_surrogate}
\end{equation}
\looseness=-1
where $\operatorname{clip}_{[l,u]}$ clips its argument to $[l,u]$, $\epsilon>0$ is the clipping threshold, and $\rho$ is the likelihood ratio between the current policy and the fixed rollout policy $\pi_{\mathrm{old}}$.
For GRPO, ratios are computed over generated tokens; environment observations only condition the policy.
Section~\ref{sec:policy_optimization} describes the integration of auxiliary step credit.
Appendix~\ref{app:actor_critic} details PPO critic targets, GAE, and complete-turn likelihood ratios.

\subsection{Prior-Data Fitted Networks}
\label{sec:pfn_background}
A prior-data fitted network (PFN) predicts the target for a query $x$ from a labeled context $\mathcal C=\{(x_j,y_j)\}_{j=1}^{m}$~\citep{PFN2021}.
For regression, $x_j,x\in\mathbb R^d$ and $y_j\in\mathbb R$.
PFNs are pretrained on synthetic supervised learning tasks to approximate posterior predictive inference under a task prior.
At deployment, a pretrained regressor evaluates
\begin{equation}
    \hat y=f_\phi(x;\mathcal C),
    \label{eq:pfn_prediction}
\end{equation}
with fixed parameters $\phi$; adapting the context changes the prediction without optimizing those parameters.
TabPFN implements this approach for tabular classification and regression~\citep{TabPFNv12022,TabPFNv2Nature2025}.
T2SPO uses this prediction interface with interaction states as queries and historical states paired with trajectory-derived labels as context.
Section~\ref{sec:method} specifies how to construct those labels and convert the resulting state predictions into auxiliary signals for policy optimization.

\section{Trajectory-to-Step Policy Optimization}
\label{sec:method}
\looseness=-1
T2SPO pairs historical states with labels derived from successful trajectories, uses these examples to predict remaining distance in subsequent rollouts, and converts adjacent-state differences into auxiliary step credit (Figure~\ref{fig:t2spo_overview}).
We describe target construction, context-based prediction, and integration into policy updates.

\subsection{Supervision from Historical Trajectories}
\label{sec:state_targets}
A completed successful trajectory provides more than a terminal score: each visited state has an observed number of turns before success.
For each successful trajectory $\tau_j$ of length $T_j$, we pair its decision state $s_{j,t}$ with the number of interactions remaining along that trajectory:
\begin{equation}
    y_{j,t}=T_j-t,\qquad t=0,\ldots,T_j-1.
    \label{eq:distance_target}
\end{equation}
Each label records the remaining length of its own successful trajectory.
These targets provide a success-conditioned distance proxy reflecting continuations observed under the collection policies.

Unsuccessful trajectories do not reveal a completion distance.
The success-only variant excludes their states from regression.
A failure-augmented variant assigns them the rollout horizon $H$ as a heuristic target and caps their share of the context.
This target is not an observed time to success; the main experiments use success-only context (Appendix~\ref{app:configurations}).

\subsection{Estimating Current States from Past Trajectories}
\label{sec:context}
\noindent\textbf{State representation.}
To use historical targets at newly visited states, we express both reference and query states in a shared feature space.
Let $h_t$ denote the instruction and interaction record available before turn $t$, including the current observation.
A frozen text encoder embeds the instruction, current observation, and the selected history from $h_t$.
The estimator's history window is configured separately from the policy context $s_t$ (Appendix~\ref{app:estimator_settings}).
We concatenate numerical step and collection-round features, then apply standardization and truncated singular value decomposition to obtain $x_t\in\mathbb R^d$.
Both transformations are fitted on the selected historical examples and shared by reference and query states.
The current action $a_t$, future observations, and the episode outcome are excluded from $x_t$.
This representation allows the estimator to compare states through their features without requiring exact state matches across trajectories.

\noindent\textbf{Shared trajectory context.}
At training round $n$, a bounded historical memory $\mathcal M_n$ supplies the labeled context $\mathcal C_n=\{(x_{j,t},y_{j,t})\}_{(j,t)\in\mathcal I_n}$.
Here $\mathcal I_n$ indexes the selected historical trajectory--turn pairs, with $|\mathcal I_n|=M_n$.
Successful state examples are sampled from completed batches and retained in a finite pool, with older entries evicted at capacity.
The failure-augmented variant additionally samples unsuccessful states subject to its quota.
A pretrained TabPFN regressor $f_\phi$ conditions on this context to predict
\begin{equation}
    \hat d_{n,t}
    = \operatorname{clip}_{[0,D_{\max}]}\bigl(f_\phi(x_t;\mathcal C_n)\bigr),
    \label{eq:distance_prediction}
\end{equation}
\looseness=-1
where $D_{\max}$ bounds the numerical output.
All batch queries share $\mathcal C_n$ and its fitted feature transformation, including adjacent states used for step credit.
Context updates refresh the labeled examples and their feature transformation, changing predictions without updating $\phi$~\citep{PFN2021,TabPFNv2Nature2025}.

\looseness=-1
\noindent\textbf{Prediction before insertion.}
We score all current rollouts using $\mathcal C_n$ before their outcomes can supply labels for $\mathcal C_{n+1}$.
This ordering lets a trajectory inform future updates while excluding its own outcome from the context used to evaluate it.
The main experiments initialize the context from a separate trajectory archive (Appendix~\ref{app:estimator_settings}).
With empty memory, training uses the base signal until sufficient historical support is available.

\subsection{From State Estimates to Step Credit}
\label{sec:progress_signal}
A remaining-distance estimate describes a state, whereas policy optimization requires a signal for the decision that changes it.
T2SPO compares estimates before and after each transition under the same historical context.
Define $\Phi_n(h_t)=-\hat d_{n,t}$ and omit $n$ below when unambiguous.
Using a potential difference~\citep{PotentialBasedRewardShaping1999}, we obtain
\begin{equation}
    F_t = \gamma\Phi_n(h_{t+1})-\Phi_n(h_t)
        = \hat d_t-\gamma\hat d_{t+1}.
    \label{eq:progress_difference}
\end{equation}
\looseness=-1
Here $\gamma\in(0,1]$ is the discount factor.
When $\gamma=1$, a positive difference indicates a reduction in predicted remaining distance.
Successful termination sets the successor potential to zero.
Unsuccessful termination receives zero final process signal, avoiding credit caused solely by ending an unsuccessful episode.
At truncation, PPO evaluates the actual successor state; GRPO sets the final process signal to zero.

To stabilize the auxiliary scale, let $z_t=\operatorname{arsinh}(F_t)$ and let $\mathcal B_n$ contain the batch turns selected by the normalization mask.
We compute the root-mean-square scale and normalized signal as
\begin{equation}
    \nu_n^2 = \frac{1}{|\mathcal B_n|}\sum_{u\in\mathcal B_n}z_u^2,
    \qquad
    \tilde F_t = \operatorname{clip}_{[-c,c]}
        \left(\frac{z_t}{\nu_n+\varepsilon}\right).
    \label{eq:progress_normalization}
\end{equation}
Here $c>0$ limits individual contributions and $\varepsilon>0$ ensures numerical stability.
The step credit supplied to the optimizer is
\begin{equation}
    p_{i,t}=w_n g_{i,t}\tilde F_{i,t},
    \label{eq:auxiliary_step_credit}
\end{equation}
\looseness=-1
where $w_n$ ramps up after burn-in and $g_{i,t}\in[0,1]$ controls whether a prediction contributes to the update.
With sufficient support, valid predictions have gate one; insufficient support or rejected predictions disable the contribution.
The ALFWorld configuration also suppresses unchanged observations and excludes them from normalization.
Turns outside the mask receive zero credit; all turns do so when $\mathcal B_n$ is empty.

Normalization, gating, and terminal exceptions modify the potential difference, so the standard policy-invariance guarantee does not extend to the full T2SPO procedure.
Algorithm~\ref{alg:step_credit} summarizes step-credit computation for the main success-only configuration.

\begin{algorithm}[t]
\caption{Computing Auxiliary Step Credit}
\label{alg:step_credit}
\begin{algorithmic}[1]
\Require Memory $\mathcal M_n$, rollouts $\mathcal T_n$, frozen encoder and regressor $f_\phi$, scale $w_n$
\Ensure Step credits $\{p_{i,t}\}$ and updated memory $\mathcal M_{n+1}$
\State Initialize $p_{i,t}\gets 0$ for all turns in $\mathcal T_n$
\State Select successful historical support states from $\mathcal M_n$
\State Assign each support state its source trajectory's remaining length (Equation~\ref{eq:distance_target})
\If{sufficient support is available}
    \State Encode support and query states; append step and collection-round features
    \State Fit the feature transformation on support; apply it to both sets
    \State Form the labeled context $\mathcal C_n$ from transformed support states
    \State Predict current and successor distances with the same $\mathcal C_n$ (Equation~\ref{eq:distance_prediction})
    \State Set successor distance to zero at successful termination
    \State Compute $F_{i,t}\gets\hat d_{i,t}-\gamma\hat d_{i,t+1}$
    \State Apply terminal and truncation rules from Section~\ref{sec:progress_signal}
    \State Determine gates $g_{i,t}$ and normalization set $\mathcal B_n$ (Section~\ref{sec:progress_signal})
    \If{$\mathcal B_n\ne\varnothing$}
        \State $z_{i,t}\gets\operatorname{arsinh}(F_{i,t})$ for $(i,t)\in\mathcal B_n$
        \State $\nu_n\gets\sqrt{|\mathcal B_n|^{-1}\sum_{(i,t)\in\mathcal B_n}z_{i,t}^2}$
        \State $p_{i,t}\gets w_n g_{i,t}\operatorname{clip}_{[-c,c]}\!\left(z_{i,t}/(\nu_n+\varepsilon)\right)$ for $(i,t)\in\mathcal B_n$
    \EndIf
\EndIf
\State Update memory with sampled successful states from $\mathcal T_n$, obtaining $\mathcal M_{n+1}$
\State \Return $\{p_{i,t}\},\mathcal M_{n+1}$
\end{algorithmic}
\end{algorithm}

\subsection{Policy Optimization with Step Credit}
\label{sec:policy_optimization}
The auxiliary term adds local information to the task-level learning signal.
We retain the base optimizer and specify how the same step credit enters its training targets.

\noindent\textbf{Group-relative optimization.}
Let $\hat A^{\mathrm{base}}_{i,t}$ be the outcome-based advantage supplied by the GRPO implementation~\citep{DeepSeekMathGRPO2024}.
T2SPO augments it at each turn:
\begin{equation}
    \hat A^{\mathrm{T2SPO}}_{i,t}
    = \hat A^{\mathrm{base}}_{i,t}+p_{i,t}.
    \label{eq:t2spo_grpo}
\end{equation}
The resulting scalar is assigned to the generated tokens of that turn and used in the token-level clipped surrogate (Equation~\ref{eq:clipped_surrogate}).
The outcome component preserves task-level supervision, while $p_{i,t}$ introduces variation among decisions within each trajectory.
We retain the base implementation's group normalization and KL regularization and introduce no learned critic.
Appendix~\ref{app:algorithm} details the multi-turn aggregation convention.

\noindent\textbf{Actor--critic extension.}
The same step credit can instead augment rewards before PPO advantage estimation:
\begin{equation}
    r^+_{i,t}=r^{\mathrm{env}}_{i,t}+p_{i,t}.
    \label{eq:t2spo_ppo}
\end{equation}
The environment reward includes task outcomes and any configured invalid-action penalty.
PPO computes advantages from these augmented rewards using GAE and trains its critic on the corresponding return targets~\citep{GAE2015,TurnPPO2025}.
This integration uses complete-turn likelihood ratios; Appendix~\ref{app:actor_critic} specifies the advantages, bootstrap handling, and objective.

The two integrations use the same source of step credit but apply it at different points: GRPO augments advantages directly, whereas PPO propagates augmented rewards through GAE.
Both keep the current batch's targets fixed during optimization and incorporate completed trajectories into subsequent contexts.
The encoder and TabPFN receive no gradient updates.

\section{Experiments}
\label{sec:experiments}

\subsection{Experimental Setup}
\label{sec:experimental_setup}
\looseness=-1
\noindent\textbf{Benchmarks and metrics.}
We evaluate on ALFWorld~\citep{ALFWorld2020} and WebShop~\citep{WebShop2022}, which require agents to complete tasks through multiple rounds of environment interaction.
For ALFWorld, we report success rates for six task categories---Pick, Look, Clean, Heat, Cool, and Pick2---and overall success.
For WebShop, we report the average task score on a 0--100 scale and the success rate.
All success rates are expressed as percentages.

\looseness=-1
\noindent\textbf{Models and baselines.}
We use Qwen2.5-1.5B-Instruct and Qwen2.5-7B-Instruct as policy models.
Our primary comparison is with GRPO~\citep{DeepSeekMathGRPO2024}, whose outcome-based training signal is augmented by T2SPO.
We also include the prompting-only base model, ReAct~\citep{ReAct2023}, PPO, RLOO, and GiGPO with standard-deviation normalization~\citep{GiGPO2025}.
All baseline results are taken from \citet{GiGPO2025}.

\noindent\textbf{Training details.}
The main T2SPO-GRPO configuration uses a context budget of 256 state examples drawn from historical successful trajectories.
We evaluate checkpoints after 150 training updates.
Training, estimator, and context settings are detailed in Appendix~\ref{app:implementation}.

\subsection{Main Results}
\label{sec:main_results}

\begin{table*}[t]
    \centering
    \caption{\textbf{Main results on ALFWorld and WebShop.}
    ALFWorld reports success rates (\%); WebShop reports score (0--100) and success rate (\%).
    RL results are means over three seeds; subscripts denote standard deviations.
    Shaded rows denote T2SPO-GRPO using only successful context examples; bold numbers indicate the highest reported mean within each model size.
    $\Delta$ reports the difference between T2SPO-GRPO and the GRPO reference within each model size: percentage points for success rates and points for WebShop score.}
    \label{tab:main_results}
    \begingroup
    \setlength{\tabcolsep}{3pt}
    \renewcommand{\arraystretch}{1.15}
    \resizebox{\textwidth}{!}{%
    \begin{tabular}{@{}ll*{7}{c}|cc@{}}
        \toprule
        Type & Method & \multicolumn{7}{c}{ALFWorld} & \multicolumn{2}{c}{WebShop} \\
        \cmidrule(lr){3-9}\cmidrule(l){10-11}
        & & Pick & Look & Clean & Heat & Cool & Pick2 & All & Score & Succ. \\
        \midrule
        \multicolumn{11}{l}{\textit{Qwen2.5-1.5B-Instruct}} \\
        Prompting & Qwen2.5 & $5.9$ & $5.5$ & $3.3$ & $9.7$ & $4.2$ & $0.0$ & $4.1$ & $23.1$ & $5.2$ \\
        Prompting & ReAct & $17.4$ & $20.5$ & $15.7$ & $6.2$ & $7.7$ & $2.0$ & $12.8$ & $40.1$ & $11.3$ \\
        RL Training & PPO (with critic) & $64.8_{\pm 3.5}$ & $40.5_{\pm 6.9}$ & $57.1_{\pm 4.9}$ & $60.6_{\pm 6.6}$ & $46.4_{\pm 4.0}$ & $47.4_{\pm 1.9}$ & $54.4_{\pm 3.1}$ & $73.8_{\pm 3.0}$ & $51.5_{\pm 2.9}$ \\
        RL Training & RLOO & $88.3_{\pm 3.0}$ & $52.8_{\pm 8.6}$ & $71.0_{\pm 5.9}$ & $62.8_{\pm 8.7}$ & $66.4_{\pm 5.5}$ & $56.9_{\pm 4.7}$ & $69.7_{\pm 2.5}$ & $73.9_{\pm 5.6}$ & $52.1_{\pm 6.7}$ \\
        RL Training & GRPO & $85.3_{\pm 1.5}$ & $53.7_{\pm 8.0}$ & $84.5_{\pm 6.8}$ & $78.2_{\pm 7.9}$ & $59.7_{\pm 5.0}$ & $53.5_{\pm 5.6}$ & $72.8_{\pm 3.6}$ & $75.8_{\pm 3.5}$ & $56.8_{\pm 3.8}$ \\
        RL Training & GiGPO & $\mathbf{94.4}_{\pm 5.9}$ & $\mathbf{67.5}_{\pm 4.6}$ & $\mathbf{94.8}_{\pm 3.8}$ & $\mathbf{94.4}_{\pm 7.8}$ & $79.8_{\pm 4.7}$ & $\mathbf{76.4}_{\pm 5.4}$ & $\mathbf{86.7}_{\pm 1.7}$ & $83.1_{\pm 1.6}$ & $65.0_{\pm 3.2}$ \\
        \rowcolor[gray]{0.93} RL Training & \textbf{T2SPO-GRPO} & $87.4_{\pm 7.4}$ & $56.8_{\pm 15.9}$ & $90.3_{\pm 10.1}$ & $65.9_{\pm 18.2}$ & $\mathbf{86.5}_{\pm 3.0}$ & $43.7_{\pm 23.9}$ & $74.7_{\pm 10.0}$ & $\mathbf{88.5}_{\pm 0.3}$ & $\mathbf{73.7}_{\pm 3.3}$ \\
        & $\Delta$ vs. GRPO & $+2.1$ & $+3.1$ & $+5.8$ & $-12.3$ & $+26.8$ & $-9.8$ & $+1.9$ & $+12.7$ & $+16.9$ \\
        \midrule
        \multicolumn{11}{l}{\textit{Qwen2.5-7B-Instruct}} \\
        Prompting & Qwen2.5 & $33.4$ & $21.6$ & $19.3$ & $6.9$ & $2.8$ & $3.2$ & $14.8$ & $26.4$ & $7.8$ \\
        Prompting & ReAct & $48.5$ & $35.4$ & $34.3$ & $13.2$ & $18.2$ & $17.6$ & $31.2$ & $46.2$ & $19.5$ \\
        RL Training & PPO (with critic) & $92.3_{\pm 4.0}$ & $64.0_{\pm 8.4}$ & $92.5_{\pm 2.4}$ & $\mathbf{89.5}_{\pm 7.0}$ & $80.3_{\pm 2.0}$ & $68.8_{\pm 8.3}$ & $80.4_{\pm 2.7}$ & $81.4_{\pm 3.1}$ & $68.7_{\pm 5.1}$ \\
        RL Training & RLOO & $87.6_{\pm 4.3}$ & $78.2_{\pm 8.3}$ & $87.3_{\pm 5.8}$ & $81.3_{\pm 7.6}$ & $71.9_{\pm 5.2}$ & $48.9_{\pm 8.4}$ & $75.5_{\pm 4.6}$ & $80.3_{\pm 3.2}$ & $65.7_{\pm 4.0}$ \\
        RL Training & GRPO & $90.8_{\pm 5.1}$ & $66.1_{\pm 6.7}$ & $89.3_{\pm 5.4}$ & $74.7_{\pm 6.9}$ & $72.5_{\pm 5.4}$ & $64.7_{\pm 7.3}$ & $77.6_{\pm 5.2}$ & $79.3_{\pm 2.8}$ & $66.1_{\pm 3.7}$ \\
        RL Training & GiGPO & $\mathbf{97.7}_{\pm 1.6}$ & $\mathbf{82.7}_{\pm 7.9}$ & $\mathbf{98.8}_{\pm 1.6}$ & $83.7_{\pm 7.2}$ & $\mathbf{89.3}_{\pm 8.2}$ & $\mathbf{79.2}_{\pm 6.6}$ & $\mathbf{90.8}_{\pm 1.3}$ & $84.4_{\pm 2.9}$ & $72.8_{\pm 3.2}$ \\
        \rowcolor[gray]{0.93} RL Training & \textbf{T2SPO-GRPO} & $91.3_{\pm 5.7}$ & $73.8_{\pm 2.1}$ & $97.3_{\pm 2.4}$ & $84.5_{\pm 16.8}$ & $78.4_{\pm 7.7}$ & $63.5_{\pm 10.5}$ & $83.6_{\pm 6.4}$ & $\mathbf{85.4}_{\pm 4.1}$ & $\mathbf{75.8}_{\pm 3.6}$ \\
        & $\Delta$ vs. GRPO & $+0.5$ & $+7.7$ & $+8.0$ & $+9.8$ & $+5.9$ & $-1.2$ & $+6.0$ & $+6.1$ & $+9.7$ \\
        \bottomrule
    \end{tabular}%
    }
    \endgroup
    \vspace{3pt}
    \begin{minipage}{\textwidth}
    \footnotesize
    Qwen2.5, ReAct, PPO, RLOO, GRPO, and GiGPO results are from \citet{GiGPO2025}.
    \end{minipage}
\end{table*}

\looseness=-1
\noindent\textbf{Comparison with baselines.}
Table~\ref{tab:main_results} compares T2SPO-GRPO with published baseline results.
Overall ALFWorld success is 74.7\% for the 1.5B model and 83.6\% for the 7B model, exceeding the reported GRPO means by 1.9 and 6.0 percentage points, respectively.
On WebShop, the corresponding differences are 16.9 and 9.7 percentage points in success rate and 12.7 and 6.1 points in task score.
T2SPO-GRPO also exceeds GiGPO in WebShop score and success at both sizes, while GiGPO retains higher overall ALFWorld success.

\noindent\textbf{Failure-augmented context.}
Table~\ref{tab:failure_context} compares success-only context with a variant that admits up to 25\% failure examples, labeled by the rollout horizon.
Both configurations use a context budget of 256 and a process weight of 0.10.
On ALFWorld, failure augmentation raises mean success from 83.6\% to 89.3\% at 7B, while the 1.5B mean remains similar at 74.0\% versus 74.7\%.
On WebShop, mean success is lower at both model sizes; task score decreases at 1.5B but increases slightly at 7B.
The effect of failure examples therefore varies across benchmarks and model sizes.
Appendix~\ref{app:failure_results} reports the ALFWorld task-category breakdown.

\begin{table}[t]
    \centering
    \caption{\textbf{Failure-augmented context on ALFWorld and WebShop.}
    Means over three seeds; subscripts report standard deviations.
    Bold numbers indicate the higher mean within each model-size pair.}
    \label{tab:failure_context}
    \begingroup
    \small
    \renewcommand{\arraystretch}{1.1}
    \setlength{\tabcolsep}{4pt}
    \begin{tabular*}{\linewidth}{@{\extracolsep{\fill}}llccc@{}}
        \toprule
        & & ALFWorld & \multicolumn{2}{c}{WebShop} \\
        \cmidrule(lr){3-3}\cmidrule(l){4-5}
        Model & Context examples & Succ. (\%) & Score & Succ. (\%) \\
        \midrule
        Qwen2.5-1.5B-Instruct & Success only & $\mathbf{74.7}_{\pm 10.0}$ & $\mathbf{88.5}_{\pm 0.3}$ & $\mathbf{73.7}_{\pm 3.3}$ \\
        & With 25\% failure examples & $74.0_{\pm 5.9}$ & $85.0_{\pm 0.6}$ & $68.2_{\pm 3.0}$ \\
        \midrule
        Qwen2.5-7B-Instruct & Success only & $83.6_{\pm 6.4}$ & $85.4_{\pm 4.1}$ & $\mathbf{75.8}_{\pm 3.6}$ \\
        & With 25\% failure examples & $\mathbf{89.3}_{\pm 3.0}$ & $\mathbf{86.3}_{\pm 0.6}$ & $74.7_{\pm 6.1}$ \\
        \bottomrule
    \end{tabular*}
    \endgroup
\end{table}

\subsection{Offline Distance Prediction}
\label{sec:offline_distance}
We evaluate remaining-distance prediction on 1,951 states from successful WebShop trajectories generated by Qwen2.5-1.5B-Instruct.
Context and query trajectories have disjoint task identities, and both use the per-trajectory targets in Equation~\ref{eq:distance_target}.
Mean absolute error (MAE) is averaged over query states and measured in turns.
Inputs contain the task description, current observation, history, step index, and pre-rollout metadata, excluding the current action and future information.

Using the same queries, we test all nine combinations of context budgets 128, 256, and 512 and SVD dimensions 32, 64, and 128.
Larger contexts retain all examples from smaller contexts.

\begin{figure}[htbp]
    \centering
    \includegraphics[width=\linewidth]{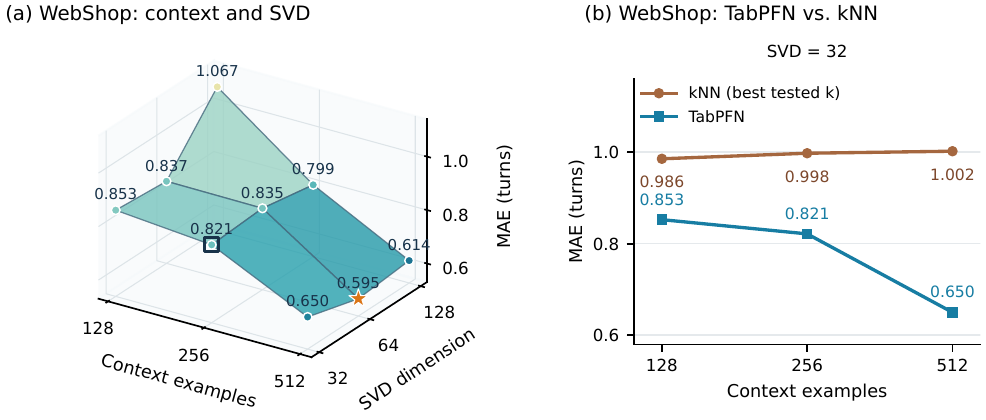}
    \caption{\textbf{Offline distance prediction on WebShop.}
    (a) Remaining-distance MAE across nine context/SVD settings; parameter axes use logarithmic spacing, and facets connect evaluated points.
    The square marks the main RL configuration; the star marks the lowest observed MAE.
    (b) TabPFN and the lowest kNN MAE across five tested neighbor counts per context budget, with matched context examples, SVD32 features, and queries.}
    \label{fig:offline_analysis}
\end{figure}

Increasing the context budget from the main RL setting of 256 to 512 reduces offline MAE at every tested SVD dimension (Figure~\ref{fig:offline_analysis}a).
At SVD32, MAE decreases from 0.8215 to 0.6500 turns, a 20.9\% reduction.
The lowest observed MAE is 0.5948 with 512 context examples and SVD64, while increasing feature dimension does not improve prediction uniformly (Appendix~\ref{app:offline_distance}).

\subsection{Comparison with Nearest Neighbors}
\label{sec:knn_comparison}
\looseness=-1
We compare TabPFN with kNN regression at SVD32 across context budgets $M\in\{128,256,512\}$.
At each budget, both estimators use the same labeled context, fitted feature transformation, and query states.
The kNN baseline averages labels of the nearest context states under cosine similarity.
We test $k\in\{8,16,M/4,M/2,M\}$ and plot the lowest observed kNN MAE at each budget (Figure~\ref{fig:offline_analysis}b).
TabPFN's MAE decreases from 0.8526 to 0.6500, while the best tested kNN MAEs are 0.9856, 0.9978, and 1.0022.
The corresponding MAE reductions are 13.5\%, 17.7\%, and 35.1\%, respectively.
Appendix~\ref{app:offline_distance} reports the full grid, tested neighbor counts, and offline evaluation protocol.

\begin{figure}[H]
    \centering
    \includegraphics[width=\linewidth]{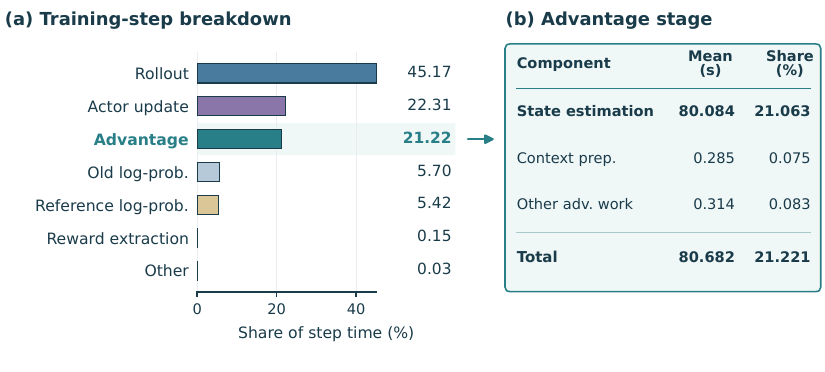}
    \caption{\textbf{Online computation cost on WebShop.}
    (a) Mutually exclusive training stages across three seeds and 360 ordinary optimizer steps.
    (b) Substages within advantage computation; times are means in seconds.
    Both panels report shares of summed full-step time.
    Other advantage work includes GRPO advantage computation, shaping, and context commit.}
    \label{fig:training_cost}
\end{figure}

\begin{samepage}
\subsection{Computation Cost}
\label{sec:computation_cost}
\looseness=-1
We profile three WebShop Qwen2.5-1.5B-Instruct runs with the main context-256/SVD32 configuration.
Across 360 optimizer steps excluding evaluation and checkpointing, mean step time is 380.21\,s.
In Figure~\ref{fig:training_cost}, rollout generation, actor updates, and advantage computation account for 45.17\%, 22.31\%, and 21.22\% of summed step time, respectively.
Within advantage computation, state estimation accounts for 21.06\% of step time, while preparing the next context takes 0.075\%.
State estimation includes embedding, context construction, SVD, and TabPFN fitting and prediction.
Appendix~\ref{app:computation_cost} details the protocol and stage timings.
\par
\end{samepage}

\par
\begin{samepage}
\section{Conclusion}
\label{sec:conclusion}
T2SPO converts historical completion lengths into auxiliary step credit through a frozen distance estimator whose context evolves with training.
Adjacent-state differences augment GRPO advantages or PPO rewards, with PPO retaining its learned critic.
T2SPO-GRPO achieves higher overall ALFWorld success and higher WebShop scores and success rates than the reported GRPO baselines at both model sizes.
In the offline WebShop evaluation, larger contexts reduce prediction error against observed remaining lengths.
At SVD32, TabPFN achieves lower MAE than the best tested kNN settings at all three context budgets.

\nopagebreak[4]
\looseness=-1
\noindent\textbf{Limitations and future work.}
The mixed effects of a fixed proportion of failure examples motivate supervision that adapts to the task and the agent's experience.
For tasks with irreversible dead ends, such as Sokoban, the estimator should distinguish recoverable detours from states where success is no longer possible.
Historical context could evolve through soft updates based on weighted combinations of feature vectors and remaining-distance targets from semantically compatible states.
When successful trajectories are scarce, memory updates and the strength of auxiliary credit could adapt to the coverage and reliability of available experience.
For more complex tasks, training or fine-tuning this compact predictor could improve task-specific progress estimation while keeping it substantially smaller than a separate LLM-based critic.
\end{samepage}

\subsection*{AI Use Statement}
AI tools assisted with manuscript editing, literature search, figure preparation, and the implementation of parts of the code.
They also supported discussions of experimental design and result interpretation.
The authors developed the core research idea, specified the algorithmic procedure, and take responsibility for the final manuscript, references, code, and reported results.

\clearpage
\appendix
\raggedbottom
\section{T2SPO Implementation and Evaluation Details}
\label{app:implementation}
This appendix details policy training, evaluation, context construction, and optimizer integration.

\subsection{Training and Evaluation Protocol}
\label{app:configurations}
Table~\ref{tab:training_configurations} summarizes the main T2SPO-GRPO training settings for the Qwen2.5-1.5B-Instruct and Qwen2.5-7B-Instruct models.
Each update collects eight trajectories for each of 16 task instances, giving 128 rollouts.
The policy receives the instruction, current observation, available actions, and up to two preceding observation--action pairs.
The estimator's history window is configured separately, as detailed in Appendix~\ref{app:estimator_settings}.

\begin{table}[htbp]
    \centering
    \caption{Policy training and evaluation settings shared by both model sizes.}
    \label{tab:training_configurations}
    \small
    \begin{tabular}{lcc}
        \toprule
        Setting & ALFWorld & WebShop \\
        \midrule
        Task instances per update & 16 & 16 \\
        Rollouts per task instance & 8 & 8 \\
        Actor learning rate & $10^{-6}$ & $10^{-6}$ \\
        KL loss coefficient & 0.01 & 0.01 \\
        Invalid-action penalty & 0.10 & 0.10 \\
        Policy history window & 2 turns & 2 turns \\
        Maximum prompt length (tokens) & 2,048 & 4,096 \\
        Maximum response length (tokens) & 512 & 512 \\
        Rollout horizon (turns) & 50 & 15 \\
        Training updates & 150 & 150 \\
        Evaluation interval (updates) & 5 & 5 \\
        Episodes per evaluation & 128 & 128 \\
        Evaluation temperature & 0.4 & 0.4 \\
        \bottomrule
    \end{tabular}
\end{table}

ALFWorld uses the environment's training and in-distribution evaluation splits.
WebShop reserves the first 500 instruction indices for evaluation and draws training tasks from the remaining indices.
Each WebShop evaluation samples 128 instructions without replacement from the reserved set.
Evaluation uses stochastic decoding at temperature 0.4, and reported results use the checkpoint after update 150.
For both the success-only and failure-augmented T2SPO-GRPO results, we report the mean and sample standard deviation over seeds 0, 1, and 2.
Baseline results in Table~\ref{tab:main_results} are taken from \citet{GiGPO2025}.

\subsection{Estimator and Context Configurations}
\label{app:estimator_settings}
\looseness=-1
Table~\ref{tab:estimator_configurations} specifies the success-only T2SPO-GRPO configurations in Table~\ref{tab:main_results}.
All use a frozen Qwen3-Embedding-0.6B encoder and the official TabPFN V2 regressor with one ensemble member and a random seed matched to the training run.
Both ALFWorld models use a structured state summary without an action suffix; WebShop uses raw observations.
The estimator includes the full available interaction history for both benchmarks.
Step and collection-round features are added before context-fitted standardization and SVD.

\begin{table}[htbp]
    \centering
    \caption{Estimator settings for the main T2SPO-GRPO runs at both model sizes. History refers to the estimator input; context and successful-state pool budgets count state rows.}
    \label{tab:estimator_configurations}
    \begingroup
    \small
    \setlength{\tabcolsep}{4pt}
    \begin{tabular}{lcc}
        \toprule
        \textbf{Shared settings} & \multicolumn{2}{c}{\textbf{Value}} \\
        \midrule
        Estimator history window & \multicolumn{2}{c}{Full prefix} \\
        TabPFN package version & \multicolumn{2}{c}{8.3.0} \\
        Shared context budget & \multicolumn{2}{c}{256} \\
        Successful-state pool capacity & \multicolumn{2}{c}{256} \\
        New successful states per round & \multicolumn{2}{c}{$\leq64$} \\
        Recent trajectory window & \multicolumn{2}{c}{256} \\
        Minimum support rows & \multicolumn{2}{c}{8} \\
        Process clip $c$ & \multicolumn{2}{c}{5} \\
        Final process weight & \multicolumn{2}{c}{0.10} \\
        Burn-in / linear ramp (rounds) & \multicolumn{2}{c}{1 / 4} \\
        Discount $\gamma$ & \multicolumn{2}{c}{1.0} \\
        \midrule
        \textbf{Benchmark-specific settings} & \textbf{ALFWorld} & \textbf{WebShop} \\
        \midrule
        State text & Structured summary & Raw \\
        SVD dimension & 64 & 32 \\
        Suppress unchanged observations & Yes & No \\
        \bottomrule
    \end{tabular}
    \endgroup
\end{table}

\looseness=-1
\noindent\textbf{Context initialization and refresh.}
Each benchmark starts with a fixed archive of 128 training trajectories collected from a separate policy checkpoint at update 100.
ALFWorld uses a separate archive for each model size, while WebShop shares one frozen archive across both sizes.
Each archive is shared across training seeds.
Successful states from this archive fill the context while the online pool is small.
Each completed training batch contributes at most 64 sampled successful states to a pool of capacity 256, with older entries evicted at capacity.
Context selection takes online states first and fills any remaining positions from the archive, so archived examples are gradually replaced as online experience accumulates.
Both sources use each state's own successful continuation length from Equation~\ref{eq:distance_target}.
The next batch shares this context and its fitted feature transformation.

\noindent\textbf{Failure-augmented context.}
The variants in Table~\ref{tab:failure_context} keep the context budget at 256 and the process weight at 0.10 on both benchmarks.
Unsuccessful states receive the horizon label $H=50$ on ALFWorld and $H=15$ on WebShop, and occupy at most 25\% of the context; successful states fill the remaining positions.
Each variant retains the corresponding success-only run's policy training and state representation settings.

\subsection{Failure-Augmented Results by Task Category}
\label{app:failure_results}
Table~\ref{tab:alfworld_failure_categories} gives the complete ALFWorld category results for the success-only and failure-augmented configurations.
At 7B, all six category means increase with failure examples; at 1.5B, the changes vary across categories.
The overall ALFWorld and WebShop comparisons are reported in Table~\ref{tab:failure_context}.

\begin{table}[htbp]
    \centering
    \caption{\textbf{Failure-augmented context across ALFWorld task categories.}
    Success rates (\%), averaged over three seeds; subscripts denote sample standard deviations.
    T2SPO-GRPO uses success-only context, while T2SPO-GRPO + Fail admits up to 25\% failure examples.
    Bold marks the higher mean at each size.}
    \label{tab:alfworld_failure_categories}
    \begingroup
    \setlength{\tabcolsep}{3pt}
    \renewcommand{\arraystretch}{1.15}
    \resizebox{\textwidth}{!}{%
    \begin{tabular}{@{}ll*{7}{c}@{}}
        \toprule
        Type & Method & \multicolumn{7}{c}{ALFWorld} \\
        \cmidrule(lr){3-9}
        & & Pick & Look & Clean & Heat & Cool & Pick2 & All \\
        \midrule
        \multicolumn{9}{l}{\textit{Qwen2.5-1.5B-Instruct}} \\
        \rowcolor[gray]{0.93} RL Training & \textbf{T2SPO-GRPO} & $\mathbf{87.4}_{\pm 7.4}$ & $56.8_{\pm 15.9}$ & $\mathbf{90.3}_{\pm 10.1}$ & $65.9_{\pm 18.2}$ & $\mathbf{86.5}_{\pm 3.0}$ & $43.7_{\pm 23.9}$ & $\mathbf{74.7}_{\pm 10.0}$ \\
        \rowcolor[gray]{0.93} RL Training & \textbf{T2SPO-GRPO + Fail} & $81.6_{\pm 4.6}$ & $\mathbf{65.2}_{\pm 17.0}$ & $82.9_{\pm 13.1}$ & $\mathbf{70.2}_{\pm 11.4}$ & $77.9_{\pm 2.6}$ & $\mathbf{54.1}_{\pm 21.2}$ & $74.0_{\pm 5.9}$ \\
        \midrule
        \multicolumn{9}{l}{\textit{Qwen2.5-7B-Instruct}} \\
        \rowcolor[gray]{0.93} RL Training & \textbf{T2SPO-GRPO} & $91.3_{\pm 5.7}$ & $73.8_{\pm 2.1}$ & $97.3_{\pm 2.4}$ & $84.5_{\pm 16.8}$ & $78.4_{\pm 7.7}$ & $63.5_{\pm 10.5}$ & $83.6_{\pm 6.4}$ \\
        \rowcolor[gray]{0.93} RL Training & \textbf{T2SPO-GRPO + Fail} & $\mathbf{94.1}_{\pm 2.9}$ & $\mathbf{75.0}_{\pm 12.5}$ & $\mathbf{98.7}_{\pm 2.2}$ & $\mathbf{87.3}_{\pm 14.5}$ & $\mathbf{83.3}_{\pm 2.9}$ & $\mathbf{82.5}_{\pm 11.0}$ & $\mathbf{89.3}_{\pm 3.0}$ \\
        \bottomrule
    \end{tabular}%
    }
    \endgroup
\end{table}

\subsection{Algorithm and Reproducibility Details}
\label{app:algorithm}
\noindent\textbf{Update order.}
For each training round, the policy first collects a rollout batch.
The estimator predicts remaining distances for the batch using only the existing context, after which the step credits and policy-training targets are computed.
The completed trajectories can then be inserted into memory and the next predictor context prepared.
This insertion does not change the predictions or targets already computed for the current batch.
The actor is optimized using these fixed targets; in the PPO extension the critic is updated as well.
There are no gradient updates to the encoder or TabPFN.

\noindent\textbf{GRPO aggregation.}
The base multi-turn implementation computes within-task outcome statistics over expanded turn rows.
Consequently, a trajectory with more turns contributes more entries to the group mean and standard deviation.
This convention differs from the standard estimator in Equation~\ref{eq:grpo_outcome}, which contributes one score per trajectory.
T2SPO preserves the base convention and adds its process term only after the base advantages have been computed.

\noindent\textbf{Validity checks.}
The direct predictor requires sufficient support, finite input features, a valid projection, and finite predictions.
Insufficient support disables the process contribution; rejected predictions receive zero gate in numerical fallback paths.
Strict error handling can instead stop an update on an estimator error.
For the ALFWorld configuration, unchanged observations set both the process gate and normalization mask to zero.
If no turns remain active, the process signal is zero.

\subsection{Actor--Critic Extension}
\label{app:actor_critic}
We detail the PPO integration from Section~\ref{sec:policy_optimization}; Appendix~\ref{app:turn_ppo_results} reports its ALFWorld results.

\noindent\textbf{Turn probabilities.}
A complete response $a_t=(a_{t,1},\ldots,a_{t,L_t})$ has autoregressive probability
\begin{equation}
    \pi_\theta(a_t \mid s_t)
    = \prod_{k=1}^{L_t}
      \pi_\theta(a_{t,k} \mid s_t, a_{t,<k}).
    \label{eq:autoregressive_policy}
\end{equation}
Environment observations condition the policy but are excluded from its action likelihood and loss.

\noindent\textbf{Advantage estimation.}
PPO uses a learned critic $V_\psi(s_t)$ to approximate the expected discounted return from $s_t$.
Generalized advantage estimation (GAE) forms discounted sums of temporal-difference residuals~\citep{GAE2015}; for a rollout segment of $K$ turns~\citep{TurnPPO2025},
\begin{equation}
    \begin{aligned}
        \delta_t
        &= r_t + \gamma V_\psi(s_{t+1}) - V_\psi(s_t), \\[2pt]
        \hat A_t^{\mathrm{GAE}}
        &= \sum_{\ell=0}^{K-t-1} (\gamma\lambda)^\ell \delta_{t+\ell},
    \end{aligned}
    \label{eq:turn_gae}
\end{equation}
where $\gamma\in(0,1]$ is the discount factor and $\lambda\in[0,1]$ controls the bias--variance tradeoff.
At a true terminal state, the successor value is zero; for a rollout truncated before termination, the final residual instead uses a bootstrapped successor value.
For T2SPO, the augmented rewards $r^+$ from Equation~\ref{eq:t2spo_ppo} replace $r_t$ in these residuals to obtain $\hat A^+$.
The critic targets are $\hat V^{\mathrm{target}}_{i,t}=\hat A^+_{i,t}+V_\psi(s_{i,t})$, formed before advantage whitening.
At truncation, the PFN also evaluates the actual successor state when constructing the auxiliary reward.

\noindent\textbf{Policy updates.}
Following turn-level optimization~\citep{TurnPPO2025}, define the likelihood ratio of the complete generated response and its clipped objective as
\begin{equation}
    \begin{aligned}
        \rho_{i,t}(\theta)
        &= \frac{\pi_\theta(a_{i,t}\mid s_{i,t})}
                 {\pi_{\mathrm{old}}(a_{i,t}\mid s_{i,t})}, \\[4pt]
        \mathcal J_{\mathrm{turn}}(\theta)
        &= \frac{1}{N_{\mathrm{tok}}}\sum_{i,t}
           \ell_{\mathrm{clip}}\bigl(\rho_{i,t}(\theta),\bar A^+_{i,t}\bigr).
    \end{aligned}
    \label{eq:t2spo_turn_objective}
\end{equation}
\looseness=-1
Here $\bar A^+$ is whitened over valid turns, and $N_{\mathrm{tok}}$ is the batch's total valid generated-token count.
Each turn contributes one clipped surrogate.
KL regularization is retained from the base optimizer.
The complete-turn log-ratio sums generated-token log-ratios, clipped to $[-20,20]$ before exponentiation for numerical stability.

\section{Offline Distance Prediction}
\label{app:offline_distance}
Table~\ref{tab:offline_distance} reports the WebShop context/SVD grid from Section~\ref{sec:offline_distance}.
The query set contains 514 trajectories and 3,073 states, with regression metrics computed on its 1,951 successful-trajectory states.
Context and query targets use each successful trajectory's own remaining length.
Tasks are split between context and query sets with seed 20260812.
A query-independent ordering of 2,115 successful context states (seed 20260920) defines nested pools of 128, 256, and 512.

All configurations encode raw observations with recent two-turn history using Qwen3-Embedding-0.6B and append the step and collection-round features.
Standardization and SVD are fitted on the selected context and then applied to the query states.
The regressor uses the official TabPFN V2 checkpoint, one ensemble member, and random seed 0.
The same cached 1,024-dimensional embeddings and held-out query states are reused for every context budget and SVD dimension.

\begin{table}[htbp]
    \centering
    \caption{\textbf{Offline remaining-distance prediction on WebShop.}
    All rows use TabPFN and evaluate the same 1,951 successful query states on held-out tasks.
    MAE and RMSE are measured in turns; $\rho$ denotes Spearman correlation.
    Bold numbers mark the best value for each metric.}
    \label{tab:offline_distance}
    \begin{tabular}{rrrrr}
        \toprule
        Context & SVD & MAE $\downarrow$ & RMSE $\downarrow$ & $\rho$ $\uparrow$ \\
        \midrule
        128 & 32 & 0.8526 & 1.4784 & 0.7293 \\
        128 & 64 & 0.8370 & 1.4714 & 0.7500 \\
        128 & 128 & 1.0666 & 1.6504 & 0.7240 \\
        \midrule
        256 & 32 & 0.8215 & 1.4698 & 0.7486 \\
        256 & 64 & 0.8345 & 1.4681 & 0.7838 \\
        256 & 128 & 0.7987 & 1.4477 & 0.7815 \\
        \midrule
        512 & 32 & 0.6500 & 1.3799 & 0.8094 \\
        512 & 64 & \textbf{0.5948} & 1.3486 & \textbf{0.8359} \\
        512 & 128 & 0.6138 & \textbf{1.3366} & 0.8333 \\
        \bottomrule
    \end{tabular}
\end{table}

\subsection{Nearest-Neighbor Comparison}
Table~\ref{tab:offline_knn} reports the full kNN comparison underlying Figure~\ref{fig:offline_analysis}b.
At each context budget, both methods use the same context examples, SVD32 feature transformation, and 1,951 successful query states.
The kNN estimates are unweighted means of labels selected by cosine similarity in the transformed feature space.
The figure uses the lowest observed kNN MAE among five tested neighbor counts at each budget: $k=32$, $64$, and $8$ for context sizes 128, 256, and 512, respectively.

\begin{table}[htbp]
    \centering
    \caption{\textbf{TabPFN and kNN on WebShop.}
    All results use SVD32; columns give context budgets $M$.
    MAE is in turns, with the lowest value in bold.
    At $k=M$, kNN returns the context mean.}
    \label{tab:offline_knn}
    \begin{tabular}{lrrr}
        \toprule
        Estimator & $M=128$ & $M=256$ & $M=512$ \\
        \midrule
        kNN ($k=8$) & 1.1042 & 1.0804 & 1.0022 \\
        kNN ($k=16$) & 1.0429 & 1.0948 & 1.0401 \\
        kNN ($k=M/4$) & 0.9856 & 0.9978 & 1.0093 \\
        kNN ($k=M/2$) & 1.0874 & 1.0882 & 1.1083 \\
        kNN ($k=M$) & 1.4719 & 1.4678 & 1.4669 \\
        \midrule
        TabPFN & \textbf{0.8526} & \textbf{0.8215} & \textbf{0.6500} \\
        \bottomrule
    \end{tabular}
\end{table}

\section{Computation Cost}
\label{app:computation_cost}
We measure online computation in the three success-only WebShop Qwen2.5-1.5B-Instruct runs, each using two NVIDIA H20 GPUs, a context budget of 256, SVD32, and a 15-turn horizon.
Each optimizer step collects 128 trajectories and queries 640--1,760 states, with a median of 928.
We retain log records from the final training process of each run and exclude steps containing evaluation or checkpoint saving, leaving 120 steps per seed and 360 steps in total.
Table~\ref{tab:training_cost_details} reports descriptive timing statistics over these steps.
Each percentage is the sum of the component's times divided by the sum of full-step times.

\begin{table}[htbp]
    \centering
    \caption{\textbf{Online training-step timing on WebShop.}
    Means and standard deviations are over 360 optimizer steps from three seeds; times are in seconds.
    Shares use summed full-step time.
    The lower block expands the advantage row in the upper block.}
    \label{tab:training_cost_details}
    \begingroup
    \small
    \setlength{\tabcolsep}{5pt}
    \renewcommand{\arraystretch}{1.12}
    \begin{tabular*}{\linewidth}{@{\extracolsep{\fill}}lrrr@{}}
        \toprule
        Component & Mean $\pm$ SD & Median & Share (\%) \\
        \midrule
        Rollout generation & $171.729\pm34.274$ & 165.729 & 45.167 \\
        Reward extraction & $0.560\pm0.215$ & 0.515 & 0.147 \\
        Old-policy log-probability & $21.657\pm6.640$ & 19.245 & 5.696 \\
        Reference-policy log-probability & $20.616\pm6.416$ & 18.093 & 5.422 \\
        Advantage computation & $80.682\pm13.747$ & 79.454 & 21.221 \\
        Actor update & $84.830\pm28.138$ & 73.509 & 22.312 \\
        Between-stage gaps & $0.131\pm0.094$ & 0.119 & 0.034 \\
        \midrule
        \textbf{Full training step} & $380.206\pm61.672$ & 363.305 & 100.000 \\
        \midrule
        \multicolumn{4}{l}{\textit{Within advantage computation}} \\
        \quad State estimation & $80.084\pm13.825$ & 78.910 & 21.063 \\
        \quad Online-context preparation & $0.285\pm0.223$ & 0.237 & 0.075 \\
        \quad Other advantage work & $0.314\pm0.134$ & 0.281 & 0.083 \\
        \bottomrule
    \end{tabular*}
    \endgroup
\end{table}

\looseness=-1
\noindent\textbf{Timing hierarchy.}
The six top-level stage timers are mutually exclusive and together cover 99.966\% of recorded step time; the residual consists of between-stage work and timer overhead.
State estimation and online-context preparation are nested within advantage computation.
The state-estimation timer covers feature and embedding handling, prediction-context construction, SVD, TabPFN fitting, and prediction.
Preparing context updates for subsequent batches is timed separately.
Together, state estimation and context preparation account for 21.138\% of step time.
The remaining advantage time includes base GRPO computation, shaping, validation, and context commit.
This runtime has no separate weight-synchronization timer, and GRPO uses no critic.
These timings describe component shares within the measured runs, rather than a matched-control slowdown.

\clearpage
\section{Additional Turn-PPO Results}
\label{app:optimizer_results}
\label{app:turn_ppo_results}
\looseness=-1
We evaluate the actor--critic integration in Appendix~\ref{app:actor_critic} on ALFWorld with both model sizes.
Our Turn-PPO control and T2SPO-Turn-PPO retain the same learned critic, GAE, and turn-level policy objective; T2SPO adds auxiliary rewards with weight 0.10.
Each update collects one rollout for each of 128 task instances, with a horizon of 50 turns and 150 training updates.
Both use actor and critic learning rates of $10^{-6}$ and $10^{-5}$, respectively, discount $\gamma=0.99$, GAE parameter $\lambda=0.9$, one PPO epoch, and clipping ratio 0.2.
The estimator uses raw state text with recent two-turn history and full admissible-action descriptions, SVD64, a context of 256 successful state examples with trajectory-local $T-t$ targets, and the official TabPFN V2 regressor from package version 2.2.1.

\begin{table}[htbp]
    \centering
    \caption{\textbf{Turn-PPO with and without T2SPO on ALFWorld.}
    Success rates (\%) are means with sample standard deviations over the same two training seeds, 0 and 1 ($n=2$).
    Each final checkpoint is evaluated on 128 episodes after 150 updates; $\Delta$ is the change in mean success in percentage points.}
    \label{tab:turn_ppo_results}
    \begingroup
    \small
    \setlength{\tabcolsep}{4pt}
    \begin{tabular*}{\linewidth}{@{\extracolsep{\fill}}lccc@{}}
        \toprule
        Model & Turn-PPO & T2SPO-Turn-PPO & $\Delta$ \\
        \midrule
        Qwen2.5-1.5B-Instruct & $91.4_{\pm 6.6}$ & $91.4_{\pm 3.3}$ & $0.0$ \\
        Qwen2.5-7B-Instruct & $90.2_{\pm 9.4}$ & $91.4_{\pm 1.1}$ & $+1.2$ \\
        \bottomrule
    \end{tabular*}
    \endgroup
\end{table}

\looseness=-1
Table~\ref{tab:turn_ppo_results} shows unchanged mean success at 1.5B and a 1.2-point increase at 7B.
At each scale, one seed improves and the other declines.

\end{document}